\documentclass[runningheads]{llncs}
\usepackage[T1]{fontenc}
\usepackage{graphicx}
\begin{document}
\title{Enhancing Robustness in Large Language Models: Prompting for Mitigating the Impact of Irrelevant  Information}
\titlerunning{Prompting for Enhancing Robustness in Large Language Models}
\author{Ming Jiang\and
Tingting Huang\and
Biao Guo \and Yao Lu\and Feng Zhang}
\authorrunning{M.Jiang et al.}
\institute{Apple
\\
\email{}} 
\maketitle              % typeset the header of the contribution
\begin{abstract}
In recent years, Large language models (LLMs) have garnered significant attention due to their superior performance in complex reasoning tasks. However, recent studies may diminish their reasoning capabilities markedly when problem descriptions contain irrelevant information, even with the use of advanced prompting techniques. To further investigate this issue, a dataset of primary school mathematics problems containing irrelevant information, named GSMIR, was constructed. Testing prominent LLMs and prompting techniques on this dataset revealed that while LLMs can identify irrelevant information, they do not effectively mitigate the interference it causes once identified. A novel automatic construction method, ATF, which enhances the ability of LLMs to identify and self-mitigate the influence of irrelevant information, is proposed to address this shortcoming. This method operates in two steps: first, analysis of irrelevant information, followed by its filtering. The ATF method, as demonstrated by experimental results, significantly improves the reasoning performance of LLMs and prompting techniques, even in the presence of irrelevant information on the GSMIR dataset.\setcounter{footnote}{0}\footnote{Dataset is available at \url{https://github.com/ymt9/GSMIR/tree/master}.}

\end{abstract}
\section{Introduction}

In recent years, Large language models (LLMs) [7, 15, 16] have demonstrated remarkable capabilities in complex reasoning tasks [12, 13, 19, 21, 25], garnering substantial interest and attention from both the research community and industry. Notably, a series of advanced prompting techniques that enhance the Chain-of-thought (COT)[2]method [3, 4, 14, 20, 22, 23, 24, 26] have significantly improved the reasoning performance of LLMs in zero-shot and few-shot scenarios.

However, the introduction of these methods does not fully reflect the true reasoning capabilities of LLMs in real-world scenarios. These methods are often bench-marked on idealised datasets where problem descriptions are closely related to the solutions. In practical applications, problem descriptions may contain information irrelevant to the task at hand. The study by Shi et al. [1] reveals that when irrelevant information is introduced into problem descriptions, the evaluated LLMs and their employed prompting techniques are all susceptible to interference from irrelevant information, leading to a significant drop in reasoning accuracy. Although Shi et al. [1] proposed combining various prompting techniques to partially mitigate the interference of irrelevant information, the highest reasoning accuracy achieved remains significantly lower than that in interference-free scenarios. This indicates that current methods still have substantial shortcomings in eliminating the impact of irrelevant information. Furthermore, Shi et al. [1] primarily focus on which types of irrelevant information LLMs are sensitive to, lacking targeted solutions.

In this paper, a mathematical problem dataset containing irrelevant information, named GSMIR and based on GSM8K [8], has been constructed to delve deeper into the fundamental reasons why LLMs are affected by irrelevant information. To the best of our knowledge, compared to the first dataset incorporating irrelevant information into GSM8K, GSMIC [1], the irrelevant information in GSMIR is more misleading. This has been done to ensure that the irrelevant information, when designed, takes into account the thematic relevance and logical connection to the original content, thereby more accurately simulating the interference that LLMs might encounter in real-world scenarios. This paper posits that the reasons LLMs are affected by irrelevant information should be attributed to two key capabilities: (1) the ability of LLMs to identify irrelevant information; (2) the ability of LLMs to self-exclude irrelevant information. In this paper, the current advanced LLMs and prompting methods have been tested on the GSMIR dataset. The experimental results indicate that LLMs possess a certain ability to identify irrelevant information, but even after identifying such information, they may still be interfered with during the reasoning process. This suggests that LLMs are not proficient at self-excluding the interference of irrelevant information.

To address this issue, this paper proposes a novel method named ATF (Analysis to Filtration Prompting). ATF primarily comprises two steps: analysis and filtration. These two steps utilise LLMs to autonomously generate demonstrations or prompts to guide the output of each step. In the analysis phase, ATF first uses prompts to guide LLMs in breaking down the input problem description into multiple clauses, then analyses each clause to determine whether it contains irrelevant information, providing reasons for its conclusions. Finally, it outputs the sentences containing irrelevant information. This process is constructed as a demonstration to guide LLMs in 
analysing new problems. In the filtration phase, ATF solely uses prompts to guide LLMs in filtering out sentences deemed to contain irrelevant information from the problem description, thereby producing a new problem description. Ultimately, LLMs use advanced prompting techniques to reason over the filtered problem. This paper employs the GPT-3 [7] series of LLMs, in conjunction with various prompting methods (Standard prompting (SP)[7], Chain-of-thought prompting (COT)[2], Zero-shot Chain-of-thought prompting (0-COT)[5], Least-to-most prompting (LTM)[6], and Instructed prompting (IP)[1] to evaluate on the GSMIR dataset. Experimental results indicate that ATF significantly improves the reasoning accuracy of LLMs when dealing with problems containing irrelevant information and is effective across all the prompting methods used in the experiments. Notably, the demonstration data employed in the identification step of ATF is sampled from other datasets. Despite this, ATF effectively enhances the rate at which LLMs identify irrelevant information. This demonstrates that LLMs do not recognise irrelevant information by learning the templates and formats from the demonstrations.

Overall, our main contributions are as follows:

1. A mathematical problem dataset containing irrelevant information has been constructed and named GSMIR. The irrelevant information in GSMIR is designed to be more thematically relevant and logically connected to the original content, better simulating real-world scenarios where problem descriptions contain irrelevant information.

2. The reasons why LLMs are affected by irrelevant information were investigated, along with their ability to identify and exclude such information. It was confirmed that, while the capability to identify irrelevant information exists in LLMs, they are incapable of self-excluding it. This inability has been posited as a primary reason for their susceptibility to irrelevant information.

3. A method named ATF has been proposed to enhance the robustness of prompt-ing methods in reasoning tasks that contain irrelevant information. ATF improves the ability of LLMs to extract and analyse problems, significantly increasing their accuracy in identifying irrelevant information. Furthermore, ATF effectively filters out any irrelevant information identified by LLMs, addressing the issue of LLMs' inability to self-exclude irrelevant information despite their identification capabilities.

\section{Related Work}
\subsection{Zero-shot Prompting}
Zero-shot prompting is a natural language processing technique that allows LLMs to leverage the knowledge acquired during their pre-training phase to handle new tasks without additional training or fine-tuning. Kojima et al. [5] discovered that when dealing with reasoning tasks, LLMs can be effectively prompted to generate logical reasoning paths and arrive at correct answers simply by inputting "Let's think step by step." Zhang et al. [4] identified a similarity-induced misleading issue in the demonstrations generated using 0-COT and found that enhancing the diversity of demonstrations can effectively mitigate this problem. However, the zero-shot prompting technique is constrained by biases in the pre-training datasets of LLMs. This study considers a more challenging scenario where the reasoning data contains irrelevant information. The sensitivity of 0-COT to such inputs was explored, and efforts were aimed at enhancing its robustness against interference from irrelevant information.
\subsection{Few-shot Prompting}
Few-shot prompting guides LLMs in addressing input questions by providing demonstrations that include input-output pairs within the prompt, thereby making full use of a small number of samples. Brown et al. [7] popularised the standard few-shot prompting by defining the demonstration format as question-answer pairs. Wei et al. [2] proposed COT prompting, which involves introducing intermediate reasoning steps in the prompt to guide LLMs in generating these steps and deducing the correct answer for the input question. Zhou et al. [6] introduced LTM (Least-to-most prompting), which, through corresponding demonstrations, instructs LLMs to decompose complex questions into a series of simpler sub-questions and solve them sequentially. Shao et al. [9] expanded the range of demonstration examples by leveraging the LLMs’ own capabilities to generate them. Several mainstream few-shot prompting techniques are explored in the context of scenarios with irrelevant information in this paper, with the finding that they are all sensitive to such information, and this issue has been effectively improved.
\subsection{The Role of Prompt Elements in Enhancing the Reasoning Capabilities of LLMs}
Recent research has begun to explore the impact of different components of prompts on the reasoning performance of LLMs. Madaan and Yazdanbakhsh [10] deconstructed prompts into three elements: symbols, text, and templates, and assessed their respective roles. Wang et al. [11] defined two core elements in the chain of thought—Bridge Object and Language Template—and confirmed their effectiveness in reasoning tasks. Shi et al. [1] investigated the impact of including irrelevant information in problem descriptions on the performance of LLMs in solving reasoning tasks, finding that the presence of irrelevant information leads to a sharp decline in reasoning performance. Building on the work of Shi et al. [1], this study designs irrelevant information that is more representative of real-world scenarios, delves deeper into the reasons behind the impact of irrelevant information on LLMs, and proposes targeted solutions to address this issue.	
\section{The GSMIR Datase}
The dataset GSMIC, introduced by Shi et al. [1], which contains irrelevant information, is effective in interfering with the CodeX series of LLMs but proves ineffective against the GPT-3.5 series of LLMs. This ineffectiveness may be attributed to: (1) the irrelevant information having little relevance to the main topic; (2) the lack of a close logical connection between the irrelevant information and the original content. The robust reasoning capabilities of LLMs enable them to easily disregard such irrelevant information. Therefore, this paper introduces the GSMIR dataset to better evaluate the ability of LLMs to resist interference from irrelevant information.
\subsection{Dataset Creation}
Each question in the GSMIR dataset includes a sentence that does not aid in solving the problem, referred to as irrelevant information. Similar to the GSMIC dataset, the GSMIR dataset randomly selects 500 data points from GSM-8K as the original problems, and then inserts a sentence containing irrelevant information into each original problem to form the new questions.
\subsection{Unrelated Information Template}
\begin{figure}
    \centering
    \includegraphics[width=0.5\linewidth]{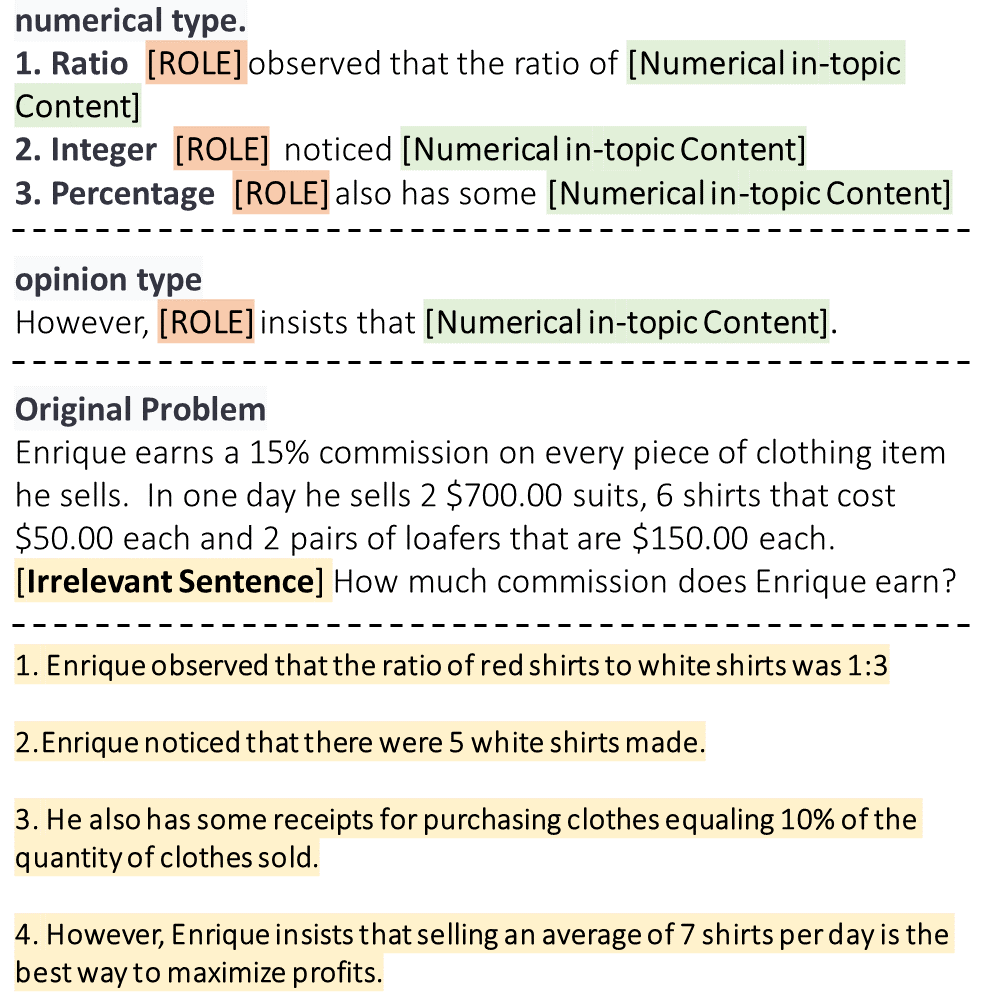}
    \caption{The key factors considered in the creation of the GSMIR dataset are presented, along with an example problem. An irrelevant sentence (highlighted in bold yellow) was inserted before the standard problem, ensuring that this sentence does not impact the derivation of the standard answer.}
    \label{fig1}
\end{figure}

To better investigate the reasoning ability of LLMs in scenarios with questions containing irrelevant information, misleading sentences that are unrelated to problem-solving have been designed. Our tests revealed that LLMs tend to focus more on numerical information during reasoning tasks. Wang et al. [17] and Liu et al. [18] demonstrated that receiving counterfactual information can lead to erroneous outputs by LLMs. Based on this, two types of irrelevant information were devised and four templates were generated. These templates (see Fig.~\ref{fig1}) were used to create the irrelevant information.

(1) Numerical Irrelevant Information:
It was observed that most problems in the GSM8K dataset contain integer values, which prompted consideration of whether LLMs are sensitive to different types of numerical information. To explore this, three distinct types of numerical irrelevant information were designed: ratios, integers, and percentages.

(2) Opinion-Based Irrelevant Information:
The impact of information containing strong subjective judgments on LLMs was aimed to be investigated. Hence, opinion-based irrelevant information was designed. For instance, given the relevant information "<Enrique sold 6 shirts>", the irrelevant information "<However, Enrique insists that selling an average of 7 shirts per day is the best way to maximise profit>" was crafted.

For [ROLE], names associated with the original question's role were selected, such as names for Role A, Role A's family members or classmates, or nouns related to the scenario. Secondly, [Numerical in-topic Content] is centred around the theme but does not aid in solving the problem.

\section{The Main Reason Why LLMs Are Affected by Irrelevant Information}
In this section, it is first investigated whether LLMs are susceptible to irrelevant information when handling reasoning tasks. Subsequently, the primary reasons behind LLMs' susceptibility to irrelevant information are analysed, and their capabilities and limitations in identifying and autonomously excluding irrelevant information are revealed. 
\subsection{How are LLMs Affected by Irrelevant Information?}
\begin{figure}
    \centering
    \includegraphics[width=0.75\linewidth]{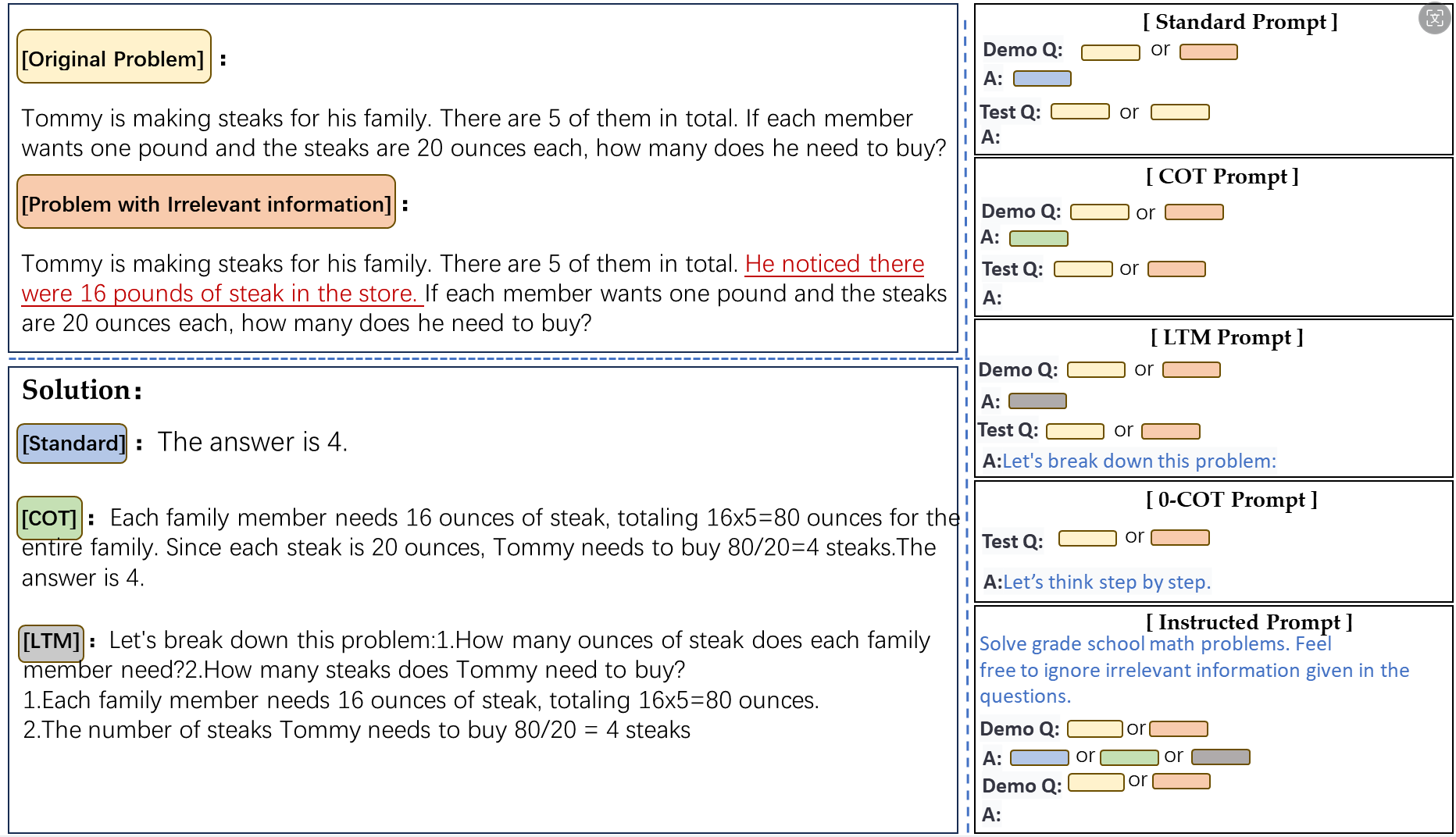}
    \caption{The various prompt formats that were employed were presented, with the use of differently coloured rectangular blocks to represent each component. The rectangular blocks on the right correspond to those of the same colour on the left (using colour coding for easier iden-tification is recommended). The "Or" symbol indicates the option to choose any one of the building blocks. [Questions with irrelevant information] are generated by adding an unrelat-ed sentence (in red font) to the [original question description].}
    \label{fig2}

\end{figure}
GPT-3.5-Turbo-16k was utilised to perform reasoning on two datasets: one be-ing the GSMIR dataset, and the other corresponding to the original questions of the GSMIR dataset, referred to in this paper as GSM8K-SLC. The accuracy of the SP and COT prompt methods on both datasets was evaluated. The design of the demon-stration set generation method is as follows: six non-overlapping data items from the GSMIC dataset were selected, and four data items were randomly drawn from the GSM8K dataset. This hybrid approach effectively stimulates LLMs' judgement capability on test questions, thereby reducing judgement bias. Unless otherwise speci-fied, the demonstration data used henceforth will adhere to the sampling strategy outlined in this section. The demonstration formats of different prompt methods are illustrated in Fig.~\ref{fig2}.
\vspace{5.0em}
\begin{table}[]
\centering
\caption{The accuracy of different prompt methods (\%) }
\label{tab1}
\begin{tabular}{cclcl}
\hline
\textbf{Method}               & \multicolumn{2}{c}{\ \ \ \ \ \ \ \ \textbf{GSM8K-SLC}} & \multicolumn{2}{c}{\ \ \ \textbf{GSMIR}} \\ \hline
SP                   & \multicolumn{2}{c}{\ \ \ \ \ \ \ \ 29.4}      & \multicolumn{2}{c}{\ \ \ 18.5}  \\
COT                  & \multicolumn{2}{c}{\ \ \ \ \ \ \ \ 77.8}      & \multicolumn{2}{c}{\ \ \ 55.2}  \\ \hline
\multicolumn{1}{l}{} & \multicolumn{1}{l}{}    &     & \multicolumn{1}{l}{}  &  
\end{tabular}

\end{table}

Table~\ref{tab1} presents the evaluation results of the two prompt methods on the two types of datasets. Compared to reasoning on the original questions of the GSM8K-SLC, the accuracy of LLMs using the two prompt methods on the GSMIR dataset decreased. This is consistent with the findings of Shi et al. [1], which indicate that the inclusion of irrelevant information in the problem description significantly reduces the reasoning accuracy of LLMs.

Our analysis suggests two potential factors contributing to the performance de-cline of LLMs: (1) LLMs lack the ability to identify irrelevant information; (2) LLMs lack the capability to exclude irrelevant information on their own. If LLMs cannot effectively identify irrelevant information, they are unable to exclude it dur-ing subsequent reasoning processes. To verify the correctness of the analysis, the experiments detailed in sections 4.2 and 4.3 were conducted.

\subsection{Can LLMs  Efficiently Identify Irrelevant Information?}
To verify whether LLMs can effectively identify irrelevant information, a straightforward validation scheme was designed. For each question in the GSMIR dataset, the inquiry was posed to the LLMs: “Does the question contain any irrelevant information? If yes, what is the irrelevant information?” The LLMs are only considered to have identified correctly if the presence of irrelevant information in the question description is accurately determined and the irrelevant information is correctly pinpointed. 

In the construction of the demonstration set, a focus was placed on guiding LLMs to output the required format rather than simply imitating the content of the demonstration. To achieve this, a prompting strategy known as Identify-Ir-Prompt was adopted, with a format that is fixed as [, Q: Does the question contain any irrelevant information? If yes, what is the irrelevant information?, A: The answer is ]. We also tested a method called Identify-Shuffle-Ir-Prompt. This method investigates whether the ability of LLMs to identify irrelevant information is influenced by the position of such information, by randomly shuffling the sentences containing irrelevant information within the demonstration questions. 

\begin{figure}
    \centering
    \includegraphics[width=0.5\linewidth]{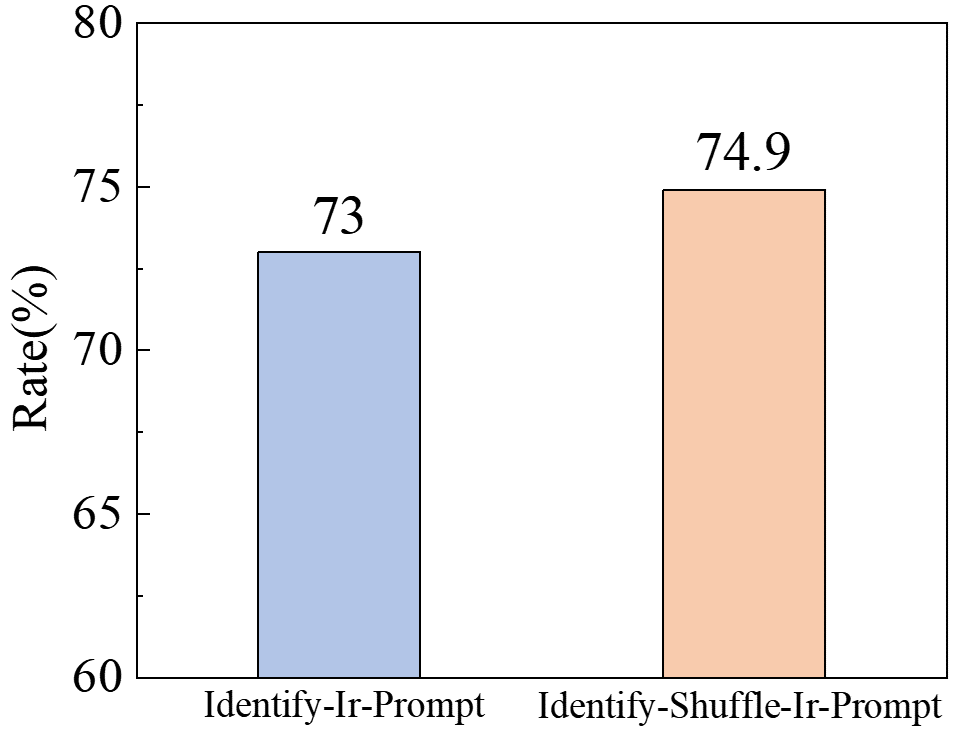}
    \caption{The identification rate of two methods for irrelevant information }
    \label{fig3}

\end{figure}
As shown in  Fig.~\ref{fig3}, the experimental results indicate that LLMs perform excellently in the task of identifying irrelevant information using both methods, with the Identify-Shuffle-Ir-Prompt method achieving an identification rate as high as 74.9\%. 

Although there is still room for improvement in the identification of irrelevant information by LLMs, this is not the primary factor causing a decline in their reasoning performance. To effectively address this issue, a deeper analysis of the challenges faced by LLMs is required. 

\subsection{Can LLMs Efficiently Exclude Irrelevant Information Themselves? }
LLMs have demonstrated a high rate of identifying irrelevant information; however, there remains a significant gap between the identification rate and the accuracy rate. This indicates that there are instances where LLMs, despite recognising irrelevant information, still provide incorrect answers. It is hypothesised that this may be due to the ineffective ability of LLMs to exclude irrelevant information. 
\begin{figure}
    \centering
    \includegraphics[width=0.5\linewidth]{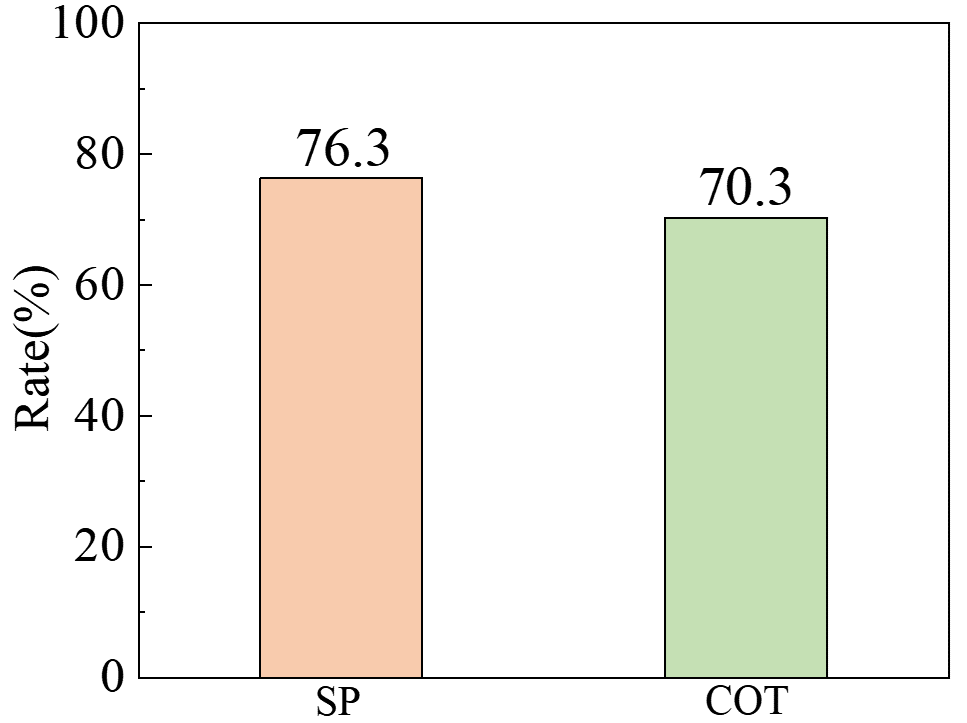}
    \caption{Statistical accuracy rates on data of errors caused by the influence of irrelevant information for both methods. }
    \label{fig4}
 
\end{figure}

To verify this hypothesis, the performance of LLMs using two prompting methods, SP and COT, on the GSMIR dataset was analysed. Initially, instances where LLMs were influenced by irrelevant information, resulting in errors, were recorded (i.e., where LLMs answered correctly on the original question but incorrectly when irrelevant information was added). With the SP method, 304 such errors were made by LLMs, while 538 errors were recorded with the COT method. Subsequently, the number of instances where LLMs correctly identified irrelevant information (SP: 232, COT: 378) was tallied and their respective proportions were calculated. As shown in Fig.~\ref{fig4}, when using the SP method, the accuracy rate for identifying irrelevant information among the erroneous data was 76.3\% (232/304). In contrast, with the COT method, this rate was 70.3\% (378/538).In the following text, we are committed to addressing this issue. 

\section{Analysis to Filtration Prompting }
\begin{figure}
    \centering
    \includegraphics[width=1\linewidth]{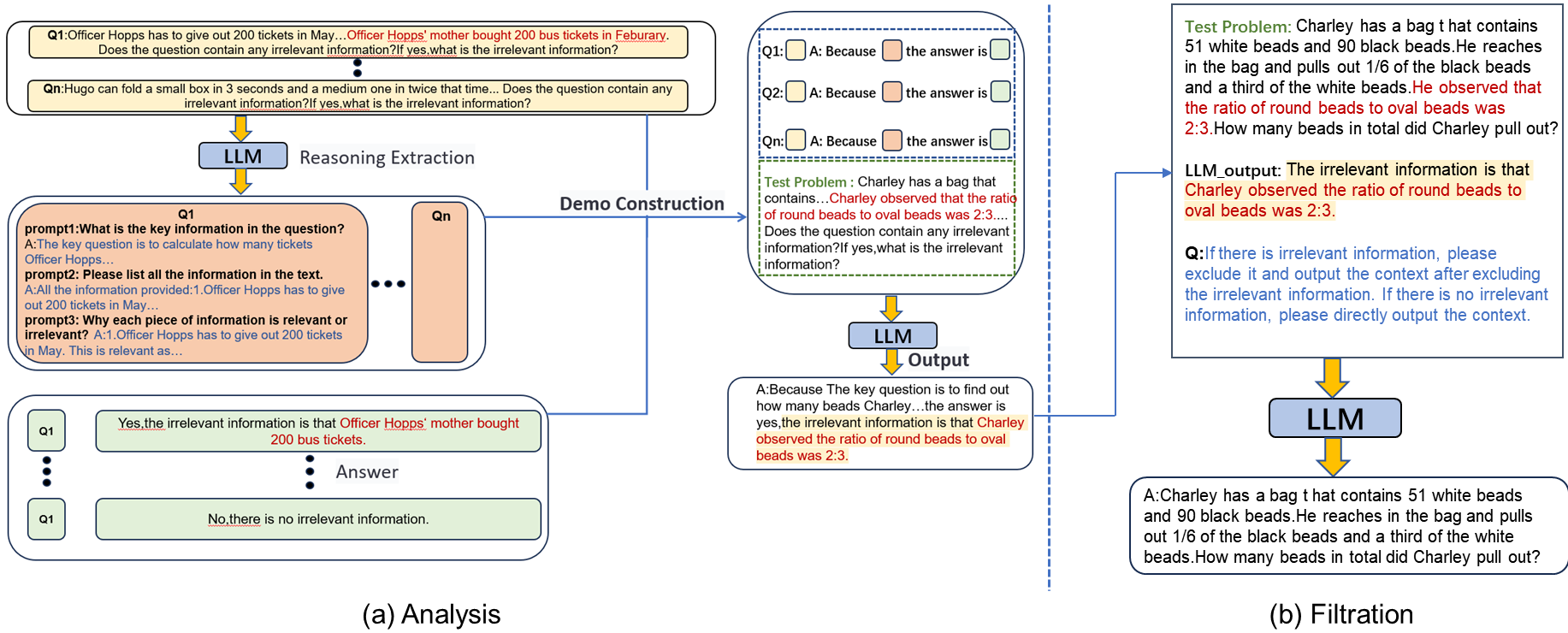}
    \caption{An overview of the ATF method includes the process of extracting information from the question, analysing the information, and identifying irrelevant information (a), as well as the process of filtering out irrelevant information from the question (b). }
    \label{fig5}
\end{figure}

A method called ATF (Analysis to Filtration Prompting) has been proposed, aimed at enhancing the robustness of LLMs against irrelevant information during the reasoning process. As illustrated in Fig.~\ref{fig5}, the ATF method primarily consists of two stages. 

\subsection{Analysis }
In the problem-solving process, the challenge is often faced that not all information is relevant to the solution. Consequently, the need to identify and focus on truly pertinent information is imperative. Based on the intuition of "analyse first, then answer," the goal of our first stage is to generate demonstrations to enhance the LLMs' ability to extract and analyse information, thereby improving the accuracy of identifying irrelevant information. To reduce the fluctuations and costs associated with manually writing demonstrations, a more efficient and stable approach was adopted—prompts were used to guide LLMs in generating demonstrations, as illustrated in Fig~\ref{fig5}. Specifically, prompts were first used to ask LLMs to identify the key information in question q, then question q was decomposed into subclauses s(i)(i=1,2...n), and prompts were used to have LLMs analyse each subclause individually, assessing whether it was irrelevant information and providing reasons. Finally, this process and the correct answer were concatenated to form a complete demonstration. 

More specifically, the demonstration d(q) includes the question q, a series of analysis processes r(q) by LLMs, and the correct answer a(q). These three parts are connected through a demonstration template, forming the structure [Q: <q>, A: Because <r(q)>, Finally, the answer is <a(q)>]. For the test question p, the input format for LLMs is [<d(q)>, Q: <p>, A: Because <r(p)>, Finally, <a(p)>], where LLMs need to output the analysis process r(p) and the answer a(p). 
\subsection{Filtration}
In the second stage, an issue was addressed where LLMs could recognise irrelevant information but were unable to exclude it, by proposing a straightforward idea—filtering out the irrelevant information identified by LLMs prior to their reasoning through the question. It was found that the irrelevant information identified by LLMs is often a highly refined and summarised version rather than complete information, which prevents the irrelevant information from being simply removed from the question. However, with the powerful comprehension capabilities of LLMs being taken into account, this issue can be resolved using prompts alone. As illustrated in Fig~\ref{fig5}, based on the answer a(p) obtained in the first stage, prompts are used to guide the model in excluding clauses containing information a(p) from the test question p, resulting in a revised question p', which needs to be solved by the LLMs. Specifically, for question p, the input format for the model is as follows: [p, a(p) Q: <If there is irrelevant information, please exclude it and output the context after excluding the irrelevant information. If there is no irrelevant information, please directly output the context.> A: Processed Context: \{p'\}], and the model needs to output the filtered p'. 
\section{Experiment }
Considering the experimental costs and effectiveness, evaluations and analyses were conducted using GPT-3.5-Turbo and GPT-3.5-Turbo-16K on the complete GSMIR dataset and the GSM8K-SLC dataset. For cost considerations, the majority of experiments were conducted on GPT-3.5-Turbo. SP,COT, 0-COT, LTM, and IP were selected as baselines. It is worth noting that both IP and ATF are prompting methods aimed at enhancing the robustness of language models against irrelevant information. Each method utilised 10 demonstrations, and the demonstration sampling strategy was consistent with Section 4.1. 
\subsection{Performance of Analysis to Filtration Prompting (ATF) on the GSMIR Dataset }

\begin{table}[]
\centering
\caption{The accuracy rates of five different prompting methods on the GSMIR dataset and the GSM8K-SLC dataset are listed, presented in the form of percentages (×100) }
\label{tab2}
\begin{tabular}{cccllc}
\hline
\textbf{Method} &  & \textbf{\ \ \ \ \ \ \ \ \ \ \ \ \ GSMIR} &  &  & \textbf{\ \ GSM8K-SLC} \\ \hline
\multicolumn{6}{c}{\textit{GPT-3.5-Turbo-16k}}                 \\ \hline
SP              &  & \ \ \ \ \ \ \ \ \ \ \ \ \ 18.5           &  &  & \ \ 29.4               \\
SP+IP           &  & \ \ \ \ \ \ \ \ \ \ \ \ \ 13.8           &  &  &                    \\
SP+ATF          &  & \textbf{\ \ \ \ \ \ \ \ \ \ \ \ \ 21.8}  &  &  &                    \\
COT             &  & \ \ \ \ \ \ \ \ \ \ \ \ \ 55.2           &  &  & \ \ 77.8               \\
COT+IP          &  & \ \ \ \ \ \ \ \ \ \ \ \ \ 43.6           &  &  &                    \\
COT+ATF         &  & \textbf{\ \ \ \ \ \ \ \ \ \ \ \ \ 64.8}  &  &  &                    \\
0-COT           &  & \ \ \ \ \ \ \ \ \ \ \ \ \ 52.4           &  &  & \ \ 77.5               \\
0-COT+IP        &  & \ \ \ \ \ \ \ \ \ \ \ \ \ 45.8           &  &  &                    \\
0-COT+ATF       &  & \textbf{\ \ \ \ \ \ \ \ \ \ \ \ \ 70.2}  &  &  &                    \\
LTM             &  & \ \ \ \ \ \ \ \ \ \ \ \ \ 56.7           &  &  & \ \ 75.9               \\
LTM+IP          &  & \ \ \ \ \ \ \ \ \ \ \ \ \ 52.4           &  &  &                    \\
LTM+ATF         &  & \textbf{\ \ \ \ \ \ \ \ \ \ \ \ \ 72.4}  &  &  &                    \\ \hline
\multicolumn{6}{c}{\textit{GPT-3.5-Turbo}}                 \\ \hline
SP              &  & \ \ \ \ \ \ \ \ \ \ \ \ \ 14.9           &  &  & \ \ 21.8               \\
SP+IP           &  & \ \ \ \ \ \ \ \ \ \ \ \ \ 11.6           &  &  &                    \\
SP+ATF          &  & \textbf{\ \ \ \ \ \ \ \ \ \ \ \ \ 18.1}  &  &  &                    \\
COT             &  & \ \ \ \ \ \ \ \ \ \ \ \ \ 50.2           &  &  & \ \ 79.3               \\
COT+IP          &  & \ \ \ \ \ \ \ \ \ \ \ \ \ 48.4           &  &  &                    \\
COT+ATF         &  & \textbf{\ \ \ \ \ \ \ \ \ \ \ \ \ 74.9}  &  &  &                    \\
0-COT           &  & \ \ \ \ \ \ \ \ \ \ \ \ \ 52.0           &  &  & \ \ 75.6               \\
0-COT+IP        &  & \ \ \ \ \ \ \ \ \ \ \ \ \ 58.9           &  &  &                    \\
0-COT+ATF       &  & \textbf{\ \ \ \ \ \ \ \ \ \ \ \ \ 71.0}  &  &  &                    \\
LTM             &  & \ \ \ \ \ \ \ \ \ \ \ \ \ 56.8           &  &  & \ \ 73.4               \\
LTM+IP          &  & \ \ \ \ \ \ \ \ \ \ \ \ \ 55.2           &  &  &                    \\
LTM+ATF         &  & \textbf{\ \ \ \ \ \ \ \ \ \ \ \ \ 69.9}  &  &  &                    \\ \hline
\end{tabular}
\end{table}

The performance of different prompting methods on the GSMIR dataset was compared in Table~\ref{tab2}. After incorporating ATF into all prompting methods, the accuracy of the models in reasoning on questions containing irrelevant information significantly improved. Notably, the most substantial performance enhancement was observed with COT, where accuracy increased from 50.2\% to 74.9\%. Conversely, the SP method saw only a modest improvement of 3.3\% in accuracy. It is important to highlight that SP's accuracy on the original questions was relatively low, at just 18.5\%. This suggests that the primary source of errors in SP during the reasoning process stemmed from computational mistakes rather than the influence of irrelevant information. Since ATF primarily enhances the robustness of prompting methods against irrelevant information rather than their computational capabilities, its effect on SP was not as pronounced. 

Interestingly, the model's accuracy decreased when using IP. It is speculated that this is because the irrelevant information in GSMIR is closely related to the question's topic.By instructing the models to ignore irrelevant information without analysis, IP inadvertently caused the loss of context relevant to the question, leading to a drop in accuracy. 

The reasoning performance of the models on the original questions was further compared with the improvements observed after the incorporation of ATF. The results indicate that, with the integration of ATF, the reasoning accuracy of the models using all prompting methods approached the accuracy on the original questions. Particularly noteworthy is the case of GPT-3.5-Turbo, where the accuracy significantly increased to 69.9\% when applying LTM combined with ATF, only 3.5\% lower than its reasoning accuracy on the original questions. 
\subsection{ATF Improves The Recognition Rate of Irrelevant Information and Robustness to Location Information}

Although detailed explanations for each identified piece of irrelevant information can be provided by Large Language Models (LLMs), there is still a need to verify whether LLMs rely on the position of irrelevant information in the demonstration to judge irrelevant information in the test questions. To this end, the ATF-Shuffle method was devised, which randomly shuffles the positions of irrelevant information in the demonstration. The recognition rates of irrelevant information by LLMs using different methods were compared, as shown in Table~\ref{tab3}. It was found that ATF-Shuffle consistently outperforms other methods in identifying irrelevant information, indicating that the capability of ATF to recognise irrelevant information is not affected by the position of irrelevant information in the demonstration. 

\begin{table}[]
\centering

\caption{the recognition rate on GSMIR(×100)}
\label{tab3}
\begin{tabular}{cc}
\hline
\textbf{Method}            & \textbf{\ \ \ \ GSMIR} \\ \hline
Identify-Ir-Prompt         & \ \ \ \ 73             \\
Identify-Shuffle-Ir-Prompt & \ \ \ \ 74.9           \\
ATF                        & \textbf{\ \ \ \ 78.9}  \\
ATF-Shuffle                & \textbf{\ \ \ \ 79.6}  \\ \hline
\end{tabular}
\end{table}

\subsection{the misjudgement rate of ATF in identifying relevant information as irrelevant is exceedingly low }

In Table~\ref{tab4}, the statistics of LLMs' recognition results using the ATF method are presented. The results indicate that the probability of LLMs identifying relevant information as irrelevant is exceedingly low, at only 2.2\%. Even in cases where LLMs fail to successfully recognise irrelevant information, they most often conclude that "no irrelevant information exists". This result suggests that even when LLMs do not successfully identify irrelevant information, they are highly unlikely to exclude information that is crucial to the reasoning process. Therefore, the ATF method largely ensures the integrity of the information necessary for reasoning. 
\begin{table}[]
\centering

\caption{ Recognition rates for the three categories of information (×100) are as follows: "irrelevant information" indicates the correct identification of irrelevant information; "other infor-mation" represents the identification of other information as irrelevant; "no irrelevant in-formation" signifies that the output contains no irrelevant information.}
\label{tab4}
\begin{tabular}{cc}
\hline
\ \ \ Irrelevant Information    & \ \ \ \ \ 0.789\ \ \ \ \ \\
\ \ \ Other Information         & \ \ \ \ \ 0.022\ \ \ \ \ \\
\ \ \ No Irrelevant Information & \ \ \ \ \ 0.189\ \ \ \ \ \\ \hline
\end{tabular}
\end{table}

\subsection{The irrelevant information that LLMs fail to recognise using ATF is often weakly irrelevant }

\begin{figure}
    \centering
    \includegraphics[width=0.5\linewidth]{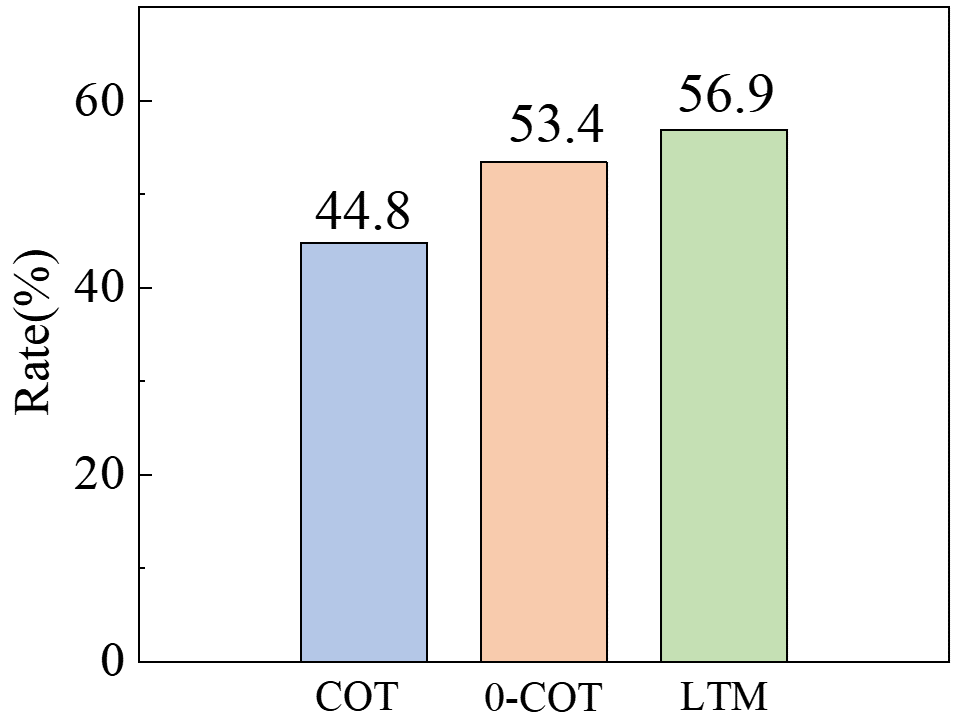}
    \caption{The proportion of weak irrelevant information in the three categories of methods. }
    \label{fig6}

\end{figure}
It has been observed that the irrelevant information which the model fails to recognise using the ATF method often constitutes "weak irrelevant information". Weak irrelevant information refers to information that, even if present in the problem description, does not affect the model's ability to deduce the correct answer. To quantify this phenomenon, experiments were conducted on the GSMIR dataset, revealing 421 instances of data where the model, when utilising the ATF method, failed to identify irrelevant information. This category of data is referred to as unrecognised problems. The data containing weak irrelevant information was statistically analysed based on the inference results from the model using three distinct prompt methods: COT, 0-COT, and LTM. The proportion of data containing weak irrelevant information within the total amount of unrecognised problems was calculated (the total amount of data containing weak irrelevant information was divided by 421), as illustrated in Fig.~\ref{fig6}. It was found that, across the three prompt methods, the proportion of unrecognised problems containing weak irrelevant information approached or even surpassed half of the total number of unrecognised problems. This finding indicates that the ATF method is significantly effective in filtering out "strong irrelevant information" that genuinely interferes with the reasoning process. 

\section{Conclusion }
In this study, the GSMIR dataset was created to explore the reasoning ability of LLMs in scenarios where problem descriptions contain irrelevant information. Additionally, the reasons why LLMs are affected by irrelevant information were investigated. Our research revealed an important phenomenon: even if an LLM can identify irrelevant information, it may still fail to exclude it autonomously. To address this issue, a method called ATF (Analysis to Filtration prompting) was designed. This method significantly enhances the ability of LLMs to recognise and filter out irrelevant information, and ATF can be combined with all existing prompting methods. Experimental results indicate that ATF effectively improves the accuracy of LLMs in reasoning on problems containing irrelevant information when using prompting methods. 

Although ATF has achieved significant success in enhancing the robustness of LLMs against irrelevant information, our current research only considers scenarios with a single piece of irrelevant information. Real-world data often contain multiple pieces of noise information, presenting a greater challenge for our method. Future researchers are encouraged to address and resolve this fundamental limitation when developing new methods and prompting techniques. Furthermore, the study of scenarios involving multiple pieces of irrelevant information and the exploration of different LLMs are left to future work.


\end{document}